\documentclass{article}

\usepackage[nonatbib,preprint]{neurips_2025}

\usepackage[utf8]{inputenc} 
\usepackage[T1]{fontenc}    
\usepackage{hyperref}       
\usepackage{url}            
\usepackage{booktabs}       
\usepackage{amsfonts}       
\usepackage{nicefrac}       
\usepackage{microtype}      
\usepackage{xcolor}         
\usepackage{graphicx}
\usepackage{amsmath}
\usepackage{amssymb}
\usepackage[square,numbers]{natbib}
\usepackage{cleveref}       
\usepackage{glossaries}
\usepackage{caption}
\usepackage{subcaption}

\makeglossaries

\title{LeDiFlow: Learned Distribution-guided\\Flow Matching to Accelerate Image Generation}

\iftrue
\author{
    Pascal Zwick$^1$, Nils Friederich$^2$, Maximilian Beichter$^2$, Lennart Hilbert$^2$,\\
    \textbf{Ralf Mikut$^2$, Oliver Bringmann$^3$}\\
    $^1$ FZI Research Center for Information Technology \hspace{4pt} $^2$ Karlsruhe Institute of Technology\\
    $3$ University of Tuebingen
}
\fi

\newacronym{afhq}{AFHQ}{Animal~Faces-HQ}
\newacronym{ae}{AE}{Autoencoder}
\newacronym{clip}{CLIP}{Contrastive Language Image Pretraining}
\newacronym{cmmd}{CMMD}{CLIP~Maximum~Mean~Discrepancy}
\newacronym{dm}{DM}{Diffusion~Model}
\newacronym{dit}{DiT}{Diffusion~Transformer}
\newacronym{ffhq}{FFHQ}{Flickr-Faces-HQ}
\newacronym{fid}{FID}{Frechet~Inception~Distance}
\newacronym{fm}{FM}{Flow~Matching}
\newacronym{fmm}{FMM}{Flow~Matching~Model}
\newacronym{gan}{GAN}{Generative~Adversarial~Network}
\newacronym{hdit}{HDiT}{Hourglass~Diffusion~Transformer}
\newacronym{kl}{KL}{Kullback-Leibler}
\newacronym{klvae}{KL-VAE}{Kullback-Leibler~Variational~Autoencoder}
\newacronym{ldm}{LDM}{Latent~Diffusion~Model}
\newacronym{lediflow}{LeDiFlow}{Learned~Distribution-guided~Flow~Matching}
\newacronym{lhq}{LHQ}{Landscapes~High-Quality}
\newacronym{ml}{ML}{Machine~Learning}
\newacronym{nn}{NN}{Neural~Network}
\newacronym{nfe}{NFE}{Numerous~Function~Evaluations}
\newacronym{ode}{ODE}{Ordinary~Differential~Equation}
\newacronym{pca}{PCA}{Principal Component Analysis}
\newacronym{sde}{SDE}{Stochastic~Differential~Equation}
\newacronym{sdv}{SDV}{Stable~Diffusion~2~VAE}
\newacronym{sota}{SOTA}{State-of-the-Art}
\newacronym{t2i}{T2I}{Text-To-Image}
\newacronym{vae}{VAE}{Variational~Autoencoder}
\newacronym{vgl}{VGL}{Variance Guided Loss}
\newacronym{vqgan}{VQ-GAN}{Vector~Quantized~Generative~Adversarial~Network}

\crefname{section}{Section}{Sections}
\creflabelformat{Section}{#2\textup{#1}#3}
\crefname{figure}{Figure}{Figures}
\creflabelformat{Figure}{#2\textup{#1}#3}
\crefname{table}{Table}{Tables}
\creflabelformat{Table}{#2\textup{#1}#3}
\crefname{equation}{Equation}{Equations}
\creflabelformat{Equation}{#2\textup{#1}#3}

\begin{document}


\maketitle
\begin{figure}[ht]
  \centering
  \includegraphics[width=1.0\textwidth]{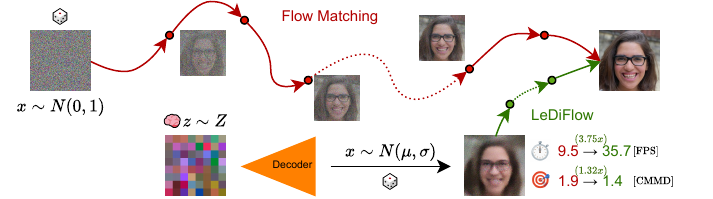}
  \caption{LeDiFlow produces an easier to solve transformation for flow matching-based generative modeling via a learned prior, enabling faster inference with higher image quality. $Z$ resembles the latent space.
  }
  \label{fig:teaser}
\end{figure}


\vspace{10pt}
\begin{abstract}

Enhancing the efficiency of high-quality image generation using \glspl{dm} is a significant challenge due to the iterative nature of the process. \gls{fm} is emerging as a powerful generative modeling paradigm based on a simulation-free training objective instead of a score-based one used in \glspl{dm}. Typical \gls{fm} approaches rely on a Gaussian distribution prior, which induces curved, conditional probability paths between the prior and target data distribution. These curved paths pose a challenge for the \gls{ode} solver, requiring a large number of inference calls to the flow prediction network. To address this issue, we present \gls{lediflow}, a novel scalable method for training \gls{fm}-based image generation models using a better-suited prior distribution learned via a regression-based auxiliary model. By initializing the \gls{ode} solver with a prior closer to the target data distribution, \gls{lediflow} enables the learning of more computationally tractable probability paths. These paths directly translate to fewer solver steps needed for high-quality image generation at inference time. Our method utilizes a \gls{sota} transformer architecture combined with latent space sampling and can be trained on a consumer workstation. We empirically demonstrate that \gls{lediflow} remarkably outperforms the respective \gls{fm} baselines. For instance, when operating directly on pixels, our model accelerates inference by up to 3.75x compared to the corresponding pixel-space baseline. Simultaneously, our latent \gls{fm} model enhances image quality on average by 1.32x in \gls{cmmd} metric against its respective baseline.

\end{abstract}


\section{Introduction}
\label{sec:intro}
Generating high-quality and diverse images efficiently represents a central and ongoing challenge in \gls{ml}. \glspl{dm}~\cite{ho2020denoising} have recently emerged as the \gls{sota} for many image synthesis tasks, surpassing earlier approaches like \glspl{gan}~\cite{goodfellow2014generative} in terms of training stability and image quality. \glspl{dm} use a diffusion process which gradually adds noise to the data in a forward process and during inference, use a noise prediction model to solve a \gls{sde} initialized with pure random noise. This approach is used in \gls{sota} text-to-image models, such as Stable~Diffusion~\cite{rombach2022high}. However, a significant operational drawback of \glspl{dm} is their iterative generation process, which generally requires many steps and thus considerable computational resources for inference. To reduce the number of iterations, Adversarial Diffusion Distillation~\cite{sauer2024add} copies principles from \glspl{gan} by using a large diffusion teacher network to train a student model, which comes at the cost of multiple training runs. \gls{fm}~\cite{lipman2022flow} presents a compelling alternative, offering advantages such as simulation-free training objectives compared to the score-based methods ~\cite{ho2020denoising,karras2022edm} typically used in \glspl{dm}. Instead of an \gls{sde}, \gls{fm} employs an \gls{ode} to define a transformation from a prior distribution $P \subseteq \mathbb{R}^d$ (e.g., a normalized Gaussian) to a target data distribution $Q \subseteq \mathbb{R}^d$ in data space $\mathbb{R}^d$. While promising, typical \gls{fm} approaches that rely on a Gaussian prior generally induce a complicated to solve, time-dependent diffeomorphic map $\phi: [0,1] \times \mathbb{R}^d \rightarrow \mathbb{R}^d$ between $P$ and $Q$. A normalized Gaussian prior lacks specific structural information about the target data distribution, forcing the model to learn a complex, non-linear transformation to map noise to intricate image features. This transformation increases the computational complexity for any numerical \gls{ode} solver, necessitating many iterations and \gls{nn} inference calls to generate high-fidelity samples.

To reduce the complexity and enhance the efficiency of \gls{fm}-based image generation, we propose \gls{lediflow}, a novel method to replace a Gaussian distribution with a learned prior distribution $P_L \subseteq \mathbb{R}^d$ via a regression-based auxiliary model (inspired by \glspl{vae}~\cite{higgins2017beta,kingma2014auto}). The core idea of \gls{lediflow} is a better-suited prior distribution $P_L$, which is learned by an auxiliary model, conditioned on a per-image latent vector, that maximizes the log-likelihood so that $P_L$ is as close to $Q$ as possible. Our method simplifies the complexity of the transformation $x \sim P_L \rightarrow y \sim Q$ that is induced by the \gls{fmm}, leading to fewer solver iterations required. The fundamental simplification of the transport problem directly translates into tangible benefits, like a reduction of the number of steps the \gls{ode} solver requires for high-quality image generation at inference time compared to standard \gls{fm} approaches with a Gaussian prior.

Unlike methods, such as Rectified~Flow~\cite{blackforestlabs2023flux,liu2022rectified}, which focuses on straightening learned paths through iterative refitting (Reflow), \gls{lediflow} simplifies the problem \textit{a priori} by defining a more advantageous starting point for the \gls{ode} solver. Our approach utilizes a scalable transformer architecture~\cite{crowson2024hourglass, dosovitskiy2020image,vaswani2017attention}, allowing for controllability through the latent space of the auxiliary model.

Our main contributions are:
\begin{itemize}
    \item We propose \gls{lediflow}, a novel method for more efficient flow matching-based image generation by using an auxiliary prior-prediction model to reduce the complexity of the transformation between a source and target distribution.
    \item We introduce a variance-guided loss for training the prior-prediction model and adapt the \gls{fm} loss with a prior based importance weighting scheme.
    \item We empirically demonstrate substantial improvements in inference speed and comparable or better image quality compared to the \gls{fm}-baseline~\cite{lipman2022flow} on \gls{sota} benchmark datasets (\gls{ffhq}~\cite{Karras2019ffhq}, \gls{lhq}~\cite{ALIS}, \gls{afhq}~\cite{choi2020stargan}) without requiring distillation or reflowing techniques.
\end{itemize}

\section{Related Work}

\label{sec:related_work}
Early approaches to image generation focused on reconstructing images from compressed representations. \glspl{ae}~\cite{hinton2006reducing} learn an encoder to map an image to a low-dimensional latent space  and a decoder to reconstruct the image, primarily aiming for dimensionality reduction~\cite{ALIS}. \glspl{vae}~\cite{higgins2017beta,kingma2014auto} extend this mapping by introducing a probabilistic latent space regularized using the \gls{kl} divergence between the learned latent distribution and a Gaussian prior~\cite{Oord2017neural}. The regularization encourages a continuous latent space suitable for generative sampling, and \gls{lediflow} adapts these principles for learning its informative prior distribution~\cite{vaswani2017attention}. \glspl{gan}~\cite{goodfellow2014generative,Oord2017neural} improve image quality by replacing the pixel-wise reconstruction loss with an adversarial loss driven by a discriminator network. This method suffers from training instabilities such as mode~collapse~\cite{Liu2019spectral} or non-convergence~\cite{Nie2019regularization} which makes \glspl{gan} harder to train than \glspl{dm}.

\subsection{Flow Matching for Generative Modeling}
\label{sec:rel_fm}

\gls{fm}~\cite{chen2018neural, lipman2022flow, liu2022rectified} offers a distinct paradigm for generative modeling. For a given prior distribution $P \subseteq \mathbb{R}^d$ and target data distribution $Q \subseteq  \mathbb{R}^d$ in data space $\mathbb{R}^d$, \gls{fm} formulates a time-dependent vector field $v : [0,1] \times \mathbb{R}^d \rightarrow\mathbb{R}^d$ that dictates the transformation of samples $x \sim P \rightarrow y \sim Q$. The transformation is called a time-dependent, diffeomorphic map $\phi : [0,1] \times \mathbb{R}^d \rightarrow\mathbb{R}^d$ and can be written as an \gls{ode}:
\begin{equation}
\begin{aligned}
    \label{eq:flow_ode}
    \frac{d}{dt} \phi_t(x) &= v_t(\phi_t(x))\\
    \phi_0(x) &= x\\
    y &= \phi_0(x) + \int_0^1 v_t(\phi_t(x)) dt,
\end{aligned}
\end{equation}
with $v_t(\phi_t(x))$ being the unknown vector value of $x$ at time $t\in[0,1]$. We define $T=1$ with $\phi_T(x) = y$ as the end-state of the transformation. This formulation is shown to be easier to solve than the score-based \gls{sde} used in \glspl{dm} \cite{lipman2022flow, liu2022rectified}. The solution to \cref{eq:flow_ode} is given by an \gls{ode} solver, which needs to access the high-dimensional vector field $v$. \cite{chen2018neural} proposes using a \gls{nn} to model the vector field. From now on, we call the network parameters $\theta$ and $f(x;\theta)$ the result when using these parameters for computation of a function $f$. The \gls{fmm} is called $\theta_\mathrm{FM}$ and is conditioned on the current timestep $t$ and $\phi_t(x) \in \mathbb{R}^d$. To train $\theta_\mathrm{FM}$, a simulation-free training objective is employed. This objective is considered 'simulation-free' as it directly regresses the target vector field from pairs of prior samples and corresponding data points, without requiring iterative simulation of a diffusion or denoising process during training. We use the rectified flow objective \cite{liu2022rectified}, which sets $\sigma_\mathrm{min} = 0$ from \cite{lipman2022flow} and can be written as:
\begin{equation}
    \label{eq:flow_loss}
    \mathcal{L}_\mathrm{CFM}(\theta_\mathrm{FM}) =  \|v_t(\phi_t(x); \theta_\mathrm{FM}) - (y - x)\|^2\\,
\end{equation}
where $\|x\|$ represents the Euclidean norm. Despite the linear training objective, the vector field $v$ induced by $\theta_\mathrm{FM}$ typically leads to a complex transformation $\phi$, that needs many iterations for accurately solving \cref{eq:flow_ode}. During training, following standard \gls{fm} practices \cite{lipman2022flow}, a random timestep $t \in [0,1]$ is sampled and a point $\phi_t(x)=yt+x(1-t)$ is calculated by linearly interpolating between sample $x \sim P$ and $y \sim Q$. This interpolation already defines the objective of $\theta_\mathrm{FM}$, learning a vector field with direct paths from $x \rightarrow y$.

\subsection{Architectures and Conditional Control}
\label{sec:rel_arch}

The performance of generative models heavily depends on the underlying network architecture. While early diffusion-based models use U-Net~\cite{karras2022edm,preechakul2022diffusion, ronneberger2015unet} style architectures, transformers~\cite{dosovitskiy2020image,vaswani2017attention} are effective in image-related tasks. The  \gls{dit}~\cite{li2022dit} adapts properties from the Vision Transformer~\cite{dosovitskiy2020image} for use in \glspl{dm}. \gls{hdit}~\cite{crowson2024hourglass} joins the hierarchical nature of U-Nets with the transformer backbone of \gls{dit}, achieving better computational efficiency when scaling to higher image resolutions. To improve the training time, \glspl{ldm}~\cite{stabilityai2024stable,rombach2022high} generates images in a smaller latent space, which then gets upsampled to a high-resolution RGB image. The latent space is induced by a \gls{vqgan}~\cite{esser2021taming,Oord2017neural}, which has been shown to have minimal impact on image quality whilst drastically reducing the image resolution and speeding up the training time of \glspl{dm}. \gls{lediflow} builds upon these advancements, utilizing a transformer-based architecture similar to \gls{hdit}~\cite{crowson2024hourglass} for both its auxiliary prior prediction and the main \gls{fmm}.

Beyond unconditional generation, controlling the output is crucial for many applications. ControlNet~\cite{zhang2023adding} demonstrates the addition of fine-grained spatial conditioning (e.g., edges, pose) to pre-trained text-to-image \glspl{dm}. Diffusion~Autoencoders~\cite{preechakul2022diffusion} combine \glspl{dm} with autoencoding principles to learn meaningful and traversable latent representations, enabling applications such as facial feature editing and manipulation. The latent space used by the auxiliary model of \gls{lediflow} also offers controllability, allowing for applications such as image interpolation, face anonymization~\cite{hukkelas2022deepprivacy2,zwick2024context} and conditional generation.

\section{LeDiFlow Architecture}
\label{sec:method}

The standard \gls{fm} process typically uses a normalized Gaussian prior $P_N = \mathcal{N}(0,1) \subseteq \mathbb{R}^d$ and a target data distribution $Q$. As mentioned earlier, training with $P_N$ induces a highly curved vector field $v$ learned by a \gls{fmm} $\theta_\mathrm{FM}$. To reduce the number of steps of the \gls{ode} solver, we propose to use a learned prior $P_L \subseteq \mathbb{R}^d$. The prior is given by an auxiliary model that is trained to predict $P_L$ as close to $Q$ as possible making the traversed vector field of the transformation computationally more tractable.

\begin{figure}
    \centering
    \includegraphics[width=1.0 \linewidth]{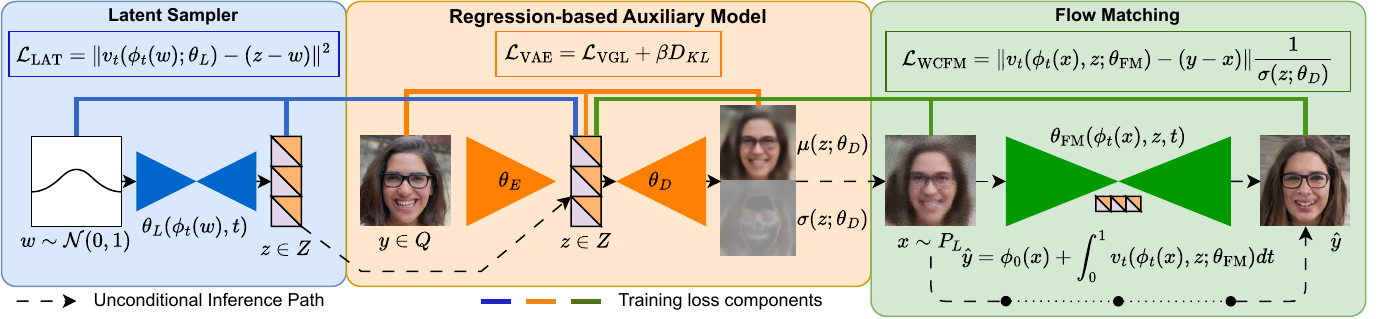}
    \caption{\gls{lediflow} Pipeline. An auxiliary model (middle) maps images \( y \sim Q \) into latent vectors \( z \sim Z \), from which the decoder defines a learned prior \( P_L = \mathcal{N}(\mu(z;\theta_D), \sigma^2(z;\theta_D)) \). In latent space (left), we train a flow model \( \theta_L \) to match trajectories from Gaussian noise \( w \sim \mathcal{N}(0,1) \) to semantic codes \( z \), using supervised vectors \( z - w \). In image space (right), another flow model \( \theta_\mathrm{FM} \) transforms samples \( x \sim P_L \) into the data distribution via flow matching towards \( y \sim Q \). Dashed lines denote forward inference, solid lines are training-time supervision.
    }
    \label{fig:fm_network}
\end{figure}

In \cref{fig:fm_network}, we give an overview of the \gls{lediflow} architecture. The training is done with three modules. First, we train an auxiliary model consisting of an encoder $\theta_E$ and decoder $\theta_D$ with the goal of obtaining a continuous latent space $Z$ and the learned prior distribution $P_L$. We define the $Z$ having a dimension of $k=256$ shaped like an image of $8 \times 8 \times 4$ pixels. Secondly, we train the \gls{fmm} $\theta_\mathrm{FM}$ conditioned on the pretrained latent space $Z$ and stochastic samples $x \sim P_L$. Last, we train a latent sampler $\theta_L$ that is used to transform a Gaussian normal distribution $N(0,1) \subseteq \mathbb{R}^k \rightarrow Z$ using traditional \gls{fm}. For image generation, we transform a sample $w \rightarrow z \in Z$, use $\theta_D$ to decode the latent vector into a per pixel distribution $P_L(z) = \mathcal{N}(\mu(z;\theta_D), \sigma(z;\theta_D))$ and draw a sample $x \sim P_L$. $x$ is then used by an \gls{ode} solver, which calls the \gls{fmm} $\theta_\mathrm{FM}$ conditioned on $z$ and $\phi_t(x)$ multiple times, to obtain a high-quality image $\hat y \in Q$. Variation is induced by the sampling process of $P_L$ and thus, a constant latent vector $z$ produces contextually similar, but visually different, results. The auxiliary model and \gls{fmm} share a similar architecture based on the \gls{hdit}~\cite{crowson2024hourglass}. The hierarchical architecture allows training on hardware with lower performance by using neighborhood attention layers instead of global ones. All models are trained individually while $\theta_\mathrm{FM}$ and $\theta_L$ require the auxiliary model to be pretrained.


\subsection{Regression-based Auxiliary Model (Distribution Prediction)}
\label{subsec:method-distr_pred}

The first module of our architecture is the auxiliary model (middle block in \cref{fig:fm_network}), which consists of the encoder $\theta_E$ and the decoder $\theta_D$. This model has two goals: The training defines the continuous latent space $Z$ where every image $y \in Q$ is mapped to a latent vector $z \sim \mathcal{N}(\mu(y;\theta_E), \sigma^2(y;\theta_E))$ using the Bayesian mapping of the auxiliary model. The second goal is training the decoder to predict the prior distribution $P_L$ to sample from for use in the \gls{fm} process. The perfect distribution would be as close to $Q$ as possible. As getting an exact solution is very challenging to do using regression training, the decoder predicts a tuple of values for every pixel, namely the mean $\mu(z;\theta_D)$ and the log variance $\log \sigma^2(z;\theta_D)$. We then define the goal as to optimize $\mu(z;\theta_D)$ and $\sigma^2(z;\theta_D)$ so that the probability of a pixel sample $y \in Q$ is as high as possible, giving a Gaussian normal distribution $\mathcal{N}(\mu(z;\theta_D), \sigma^2(z;\theta_D))$:
\begin{equation}
    \mathcal{N}(y | \mu(z;\theta_D), \sigma^2(z;\theta_D)) = \frac{1}{\sqrt{2\pi\sigma^2(z;\theta_D)}} e^{-\frac{(y - \mu(z;\theta_D))^2}{2\sigma^2(z;\theta_D)}}.
\end{equation}
Optimizing the Gaussian distribution is complicated as the optimization target function is flat far away from the optimum, which leads to vanishing gradients at large $|y - \mu(z;\theta_D)|$. For this reason, we optimize $\log \mathcal{N}(\mu(z;\theta_D), \sigma^2(z;\theta_D))$, which has the same minimum as the normal distribution itself. This process is similar to the optimization of the log-likelihood function \cite{conniffe1987}. We call our loss the \gls{vgl}:
\begin{equation}
\begin{aligned}
    \mathcal{L}_\text{VGL}(y | \mu(z;\theta_D)), \sigma(z;\theta_D)) &= -\log \mathcal{N}(y | \mu(z;\theta_D), \sigma(z;\theta_D))
    \\ &= \frac{1}{2} \left( \frac{(y - \mu(z;\theta_D))^2}{ \sigma^2(z;\theta_D)} + \log \sigma^2(z;\theta_D) + \log(2 \pi) \right).
\end{aligned}
\end{equation}
During training, the log-variance $\log \left( \sigma^2(z;\theta_D) \right)$ is predicted by the \gls{nn}. The decoder parameters $\mu(z;\theta_D)$ and $\sigma(z; \theta_D)$ define a learned prior distribution $P_L = \mathcal{N}(\mu(z;\theta_D), \sigma(z, \theta_D))$, from which we draw initial image samples $x \sim P_L$. These serve as the starting point for the \gls{fm} process. We then combine the reconstruction and \gls{kl} loss the following way:
\begin{equation}
    \mathcal{L}_\mathrm{VAE}(\theta_E, \theta_D) = \beta D_{KL}(Z || \mathcal{N}(0,1);\theta_E) + \mathcal{L}_\mathrm{VGL}(\theta_D),
\end{equation}
where $\beta$ is a user controlled parameter \cite{higgins2017beta}. A lower $\beta$ leads to better reconstruction but a worse feature mapping in latent space. We found $\beta = 10^{-3}$ to be a good fit.

\subsection{Flow Matching}
\label{subsec:method-fmm}
The second module of our architecture is the \gls{fmm} $\theta_\text{FM}$ used by the \gls{ode} solver for transporting samples $x \sim P_L$ to image values $y \sim Q$. The \gls{fmm} is based on an \gls{hdit}~\cite{crowson2024hourglass} and consists of multiple neighborhood and global attention layers, down- / upsample layers and skip connections. We train the model by decoding images $y \in Q$ from the dataset using $\theta_E$ into a latent representation $z = (\mu(y;\theta_E), \sigma(y;\theta_E)) \in Z$ and an initial image distribution $P_L$. We then sample $\hat z \sim \mathcal{N}(\mu(y;\theta_E), \sigma^2(y;\theta_E))$ and $x \sim P_L$. The model $\theta_\text{FM}$ is then conditioned on $\hat z$ and a random timestep $t \in [0,1]$ is used for the training sample $\phi_t(x)$. The \gls{fm} training is then done similar to the standard flow matching procedure \cite{lipman2022flow} with one key difference in our approach being the introduction of an importance scaling term, $\frac{1}{\sigma(z;\theta_D)}$, applied per pixel to the squared error in the loss function.
Here, $\sigma(z;\theta_D)$ represents the standard deviation of our learned prior distribution $P_L$.
This weighting strategy serves a dual purpose:
\begin{enumerate}
    \item \textbf{Refining confident predictions:} For pixels where $\mu(z;\theta_D)$ is confident (i.e., $\sigma(z;\theta_D)$ is small), the term $\frac{1}{\sigma(z;\theta_D)}$ becomes large, which up-weights the loss for these pixels. The weighting encourages $\theta_\mathrm{FM}$ to learn very precise adjustments effectively refining the already strong predictions from $\theta_D$. Dividing by $\sigma(z;\theta_D)$ can generate large weight values, which we prevent by clipping the divider to a maximum value of $10^{-3}$. This way, bad pixels with a very large weight do not add to the loss relative to the rest of the image pixels and batch.
    \item \textbf{Stabilizing learning for uncertain predictions:} Conversely, for pixels where the prior is uncertain (i.e., $\sigma(z;\theta_D)$ is large), the loss contribution is down-weighted, which prevents noisy pixels (i.e. difficult to estimate) from dominating the training signal of $\theta_\mathrm{FM}$, potentially leading to a more stable vector field $v$.
\end{enumerate}
The importance-weighted loss is given by:
\begin{equation}
\label{eq:flow_loss_importance}
    \mathcal{L}_\mathrm{WCFM}(\theta_\mathrm{FM}) = \| v_t(\phi_t(x), \hat z;\theta_\mathrm{FM}) - (y - x) \|^2 \frac{1}{\sigma(z;\theta_D)}.
\end{equation}


\subsection{Latent Sampling}
\label{subsec:method-lat_samp}
To generate new randomized data samples $\hat y \in Q$, we need to be able to sample the latent space $Z$. As we use the \gls{kl}-Divergence for regularization, the latent space learned by our auxiliary model should be continuous. To sample $z \sim Z$, we need to either do a \gls{pca}~\cite{pearson1901pca} or transform from an easier-to-sample distribution to the latent one. As PCA does not work well for arbitrary distributions, we opted for the second solution, also used similarly in diffusion autoencoders~\cite{preechakul2022diffusion}. Our latent space $Z$ has the shape $8 \times 8 \times 4$ and we can treat it like an image with four channels. We then use a \gls{fmm} $\theta_L$, which is built on a simplified \gls{hdit} architecture, and the standard \gls{fm} training method from \cref{sec:rel_fm}. The loss from \cref{eq:flow_loss} is used to transform a normalized Gaussian distribution $P_N \rightarrow Z$. The latent sampler is placed on the left in \cref{fig:fm_network} and uses the latent encodings $z \in Z$ calculated by the pretrained encoder model $\theta_E$ for its training loop.

\section{Results}
\label{sec:results}

In this section, we show results using models trained on a consumer workstation (Intel Core i7-10700K, 64GB RAM, two NVIDIA RTX 3090) and  a server (76 CPU cores, four NVIDIA A100 40GB GPUs). The server handles the more compute and memory intensive models for operation in RGB pixel space explained in \cite{crowson2024hourglass}. The workstation handled training with pre-encoded images using the pretrained \gls{sdv}. This encoder compresses RGB images into a latent space representation with an image size reduction of 8 whilst increasing the channel count to 4. Thus, images with the size of $256^2 \times 3$ are compressed into an image with size $32^2 \times 4$. All models were trained in approximately \~30h. We compare our method to the reference \gls{fm} implementation using the same \gls{hdit} architecture in unconditional and conditional generation tasks. We evaluate our findings using the \gls{ffhq}, \gls{afhq}, \gls{lhq} and ImageNet datasets.

\subsection{Unconditional Generation}
\label{subsec:results-uncond_generation}

\begin{table}
    \centering
    \caption{Performance and quality comparison of the \gls{hdit}-S/4 \cite{crowson2024hourglass} architecture using the standard \gls{fm} \cite{lipman2022flow} algorithm and the method proposed in this work using different datasets at $\boldsymbol{256^2}$ resolution. The performance is measured on an RTX 4070 Super with batch size 64. \gls{fm} uses 8 inference steps while \gls{lediflow} uses 4 latent + 2 \gls{fm} steps for the Midpoint method and 2 + 1 steps for the 3rd-order Heun method (Heun3 \cite{Butcher1967}). $\dagger$ The decoding cost using the \gls{sdv} decoder is $\approx 20 \mathrm{ms}$ per image.}
    \begin{tabular}{ccccc}
        \hline
        \textbf{Method} & \gls{fm} & \textbf{\gls{lediflow}} & \gls{fm} (\gls{sdv}) & \textbf{\gls{lediflow} (\gls{sdv})} \\
        \# Parameters & 160M& 213M & 154M  & 207M \\
        \hline
        \textbf{Evaluation (CMMD)} $\downarrow$  & & & & \\
        \gls{ffhq} \cite{Karras2019ffhq} & 2.48 & 2.66 & 2.00 & \textbf{1.08} \\
        \gls{afhq} \cite{ALIS} & 1.08 & 1.33 & \textbf{0.81} & 0.87 \\
        \gls{lhq} \cite{choi2020stargan} & 1.55 & 1.39 & 1.48 & \textbf{1.21} \\
        ImageNet \cite{deng2009imagenet} & - & - & 3.10 & \textbf{2.45}\\
        \hline
        \textbf{Inf. time [ms/img]} $\downarrow$ \\
        Midpoint & 105 & \textbf{32} & 80$^\dagger$ & 40$^\dagger$ \\
        Heun3~\cite{Butcher1967} & 116 & \textbf{28} & 85$^\dagger$ & 37$^\dagger$ \\
        \hline
    \end{tabular}
    \label{tab:inf_performance}
\end{table}

\begin{figure}
    \centering
    \includegraphics[width=0.32\linewidth]{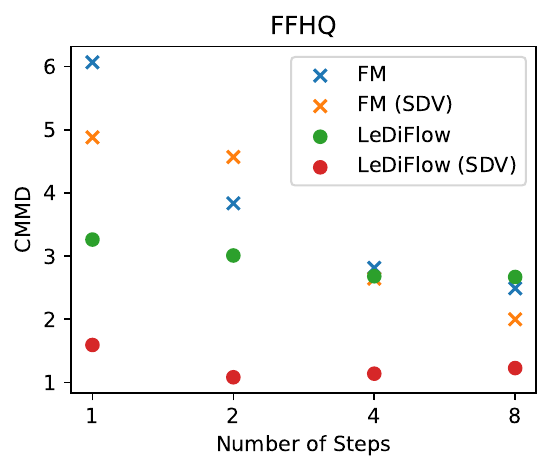}
    \includegraphics[width=0.32\linewidth]{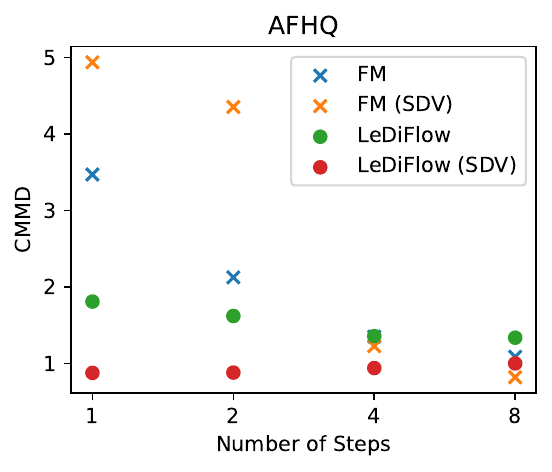}
    \includegraphics[width=0.33\linewidth]{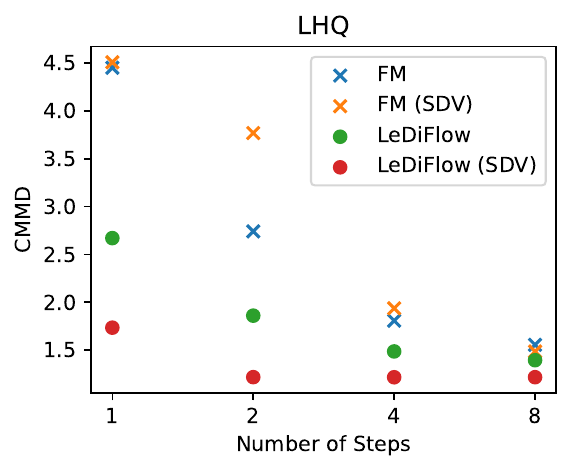}
    \caption{Visual scatter plot showing the comparison from \cref{tab:inf_performance}.}
    \label{fig:inf_performance}
\end{figure}

\begin{figure}
    \centering
    \includegraphics[width=0.3\linewidth]{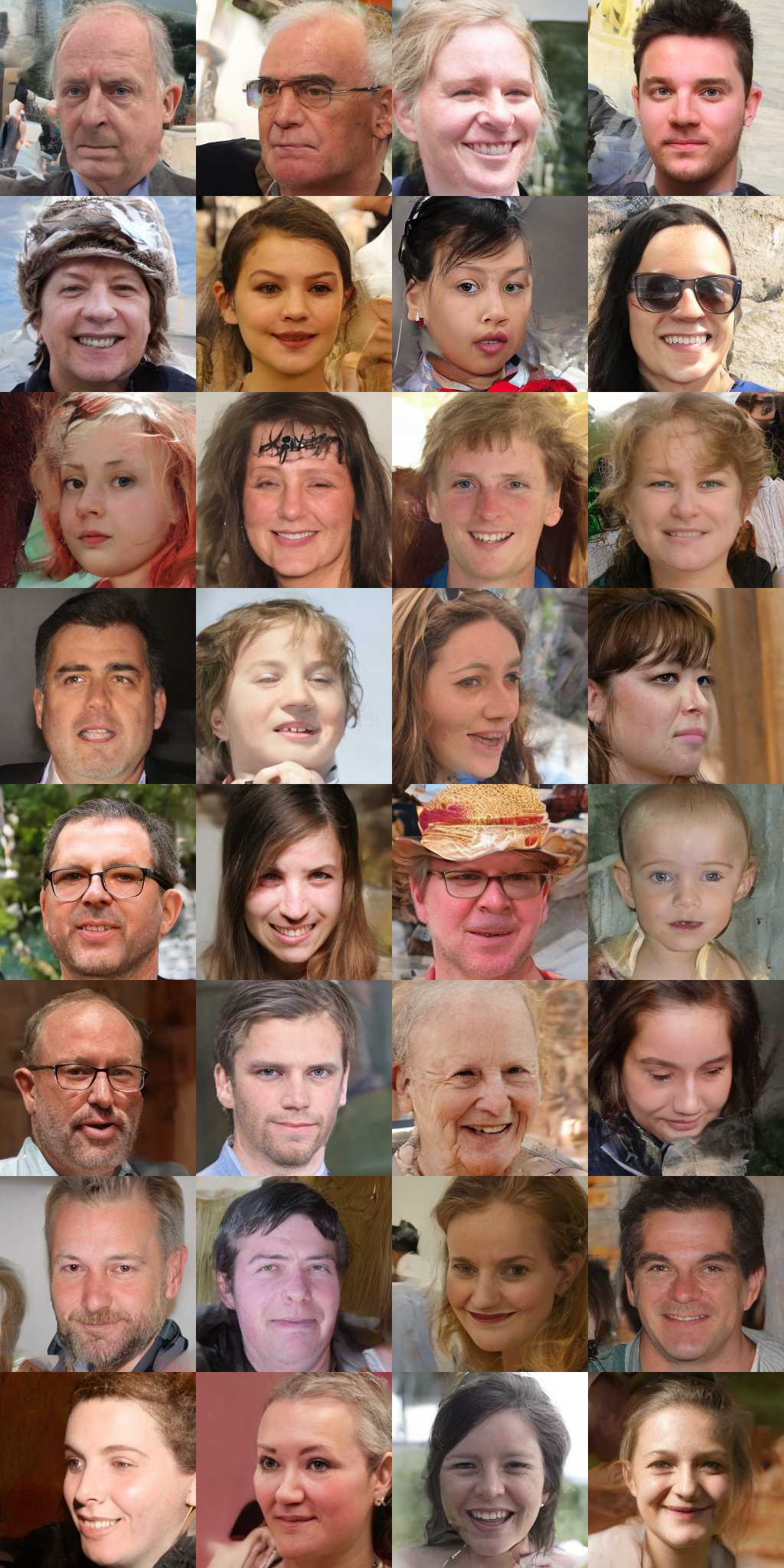}
    \includegraphics[width=0.3\linewidth]{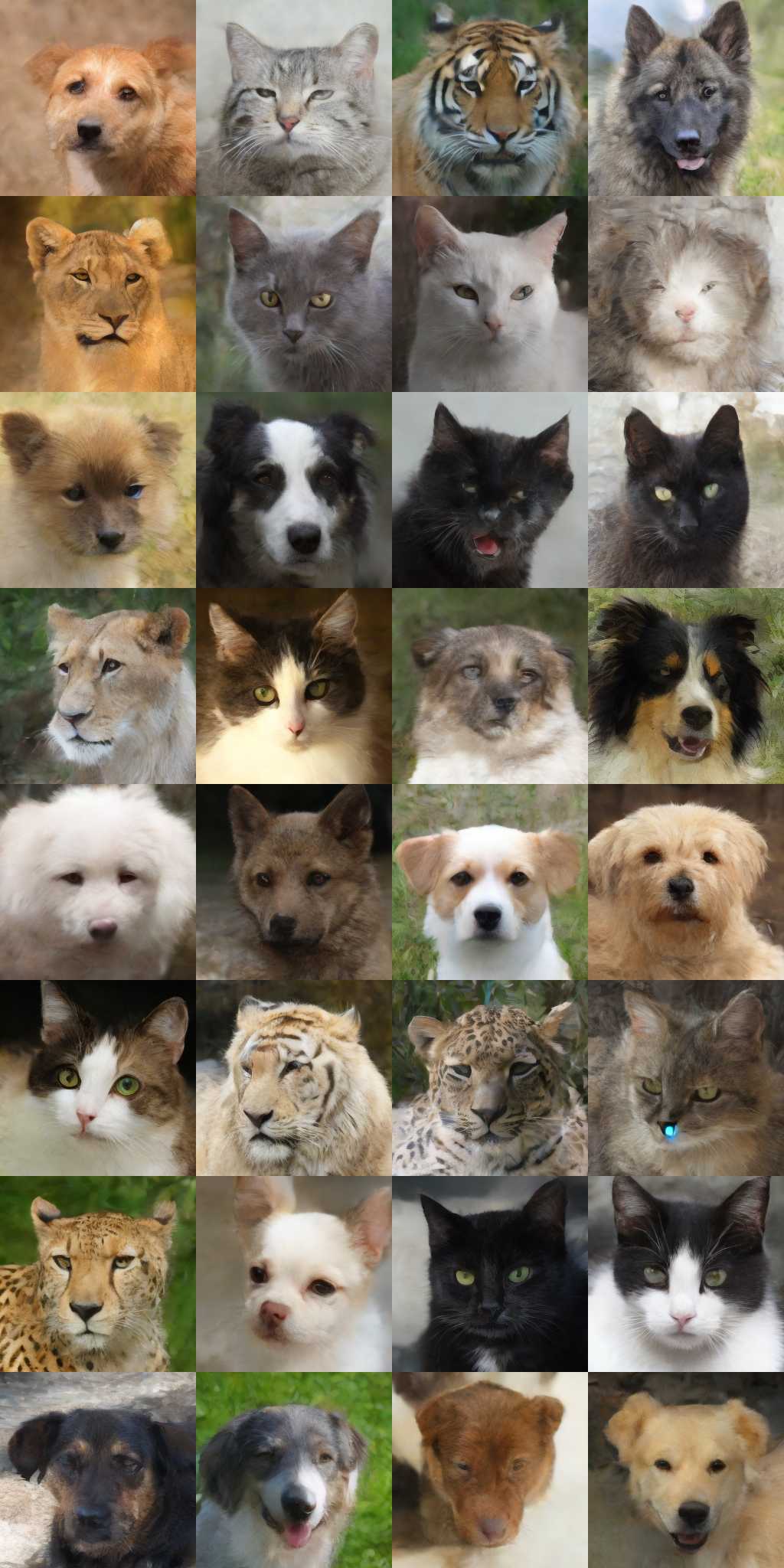}
    \includegraphics[width=0.3\linewidth]{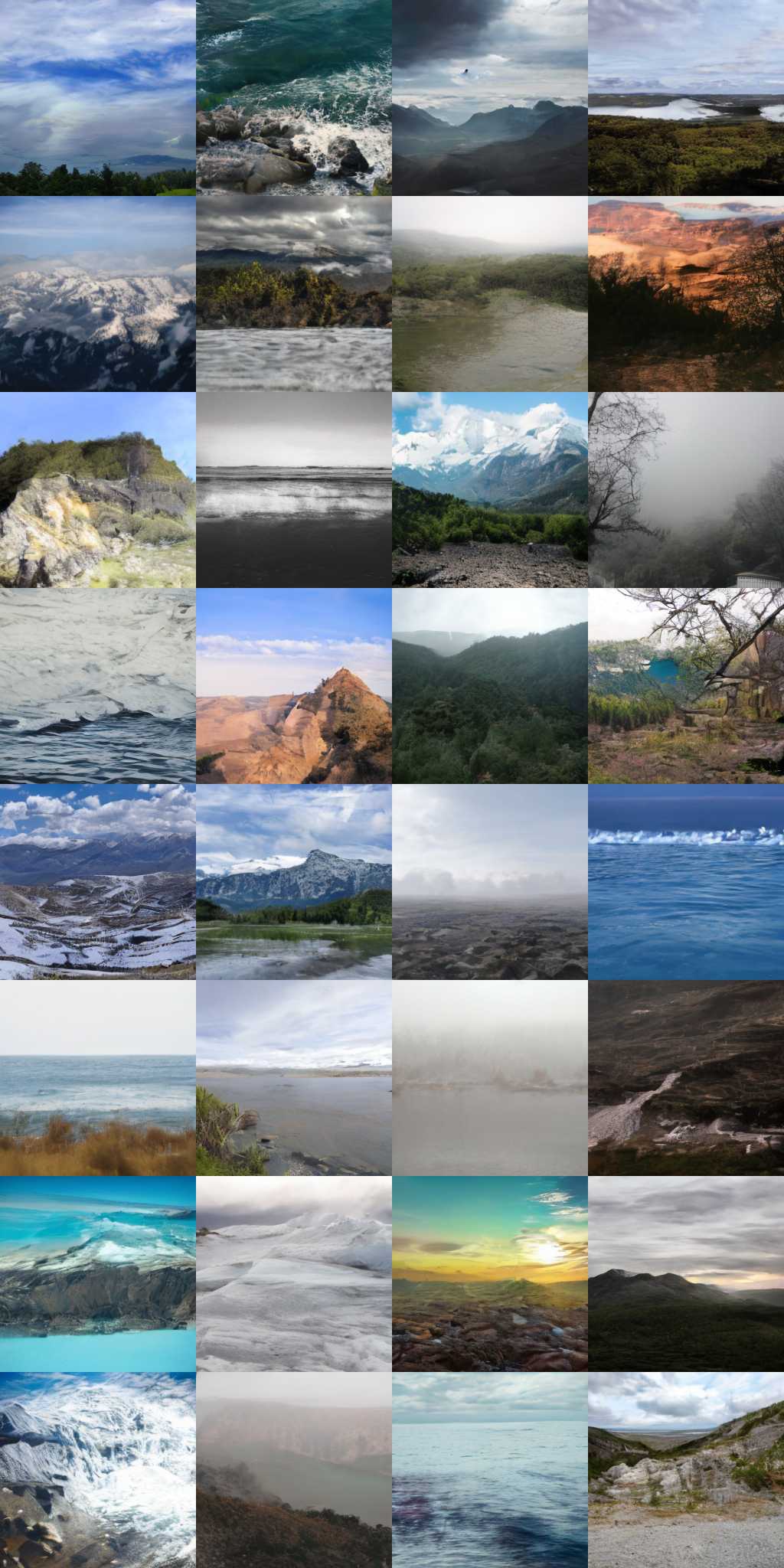}
    \caption{Samples showing unconditional generation of our method using 4 inference steps (Midpoint solver). From left to right: \gls{ffhq}~\cite{Karras2019ffhq}, \gls{afhq}~\cite{ALIS}, \gls{lhq}~\cite{choi2020stargan}}
    \label{fig:generation_unconditional}
\end{figure}

In \cref{tab:inf_performance} and \cref{fig:inf_performance}, we compare the methods in quality and inference time. The quality is assessed using the \gls{cmmd}~\cite{Jayasumana2023rethinking} metric. \gls{cmmd} was chosen because it is a more recent image quality metric and more stable compared to the \gls{fid} ~\cite{heusel2017gans}. The results show that our method achieves speed improvements in inference while maintaining competitive or superior image quality. The scatter plots in \cref{fig:inf_performance} show that our method converges faster and produces comparable results with only one step of flow matching using an \gls{ode} solver of 2nd order or higher (i.e. midpoint, Heun). With \gls{lediflow} using an auxiliary prior prediction model, the cost of initializing the \gls{ode} solver is higher compared to using $P_N$. Despite this cost, our method up to 3.75x faster than the default \gls{fm} method, as \gls{lediflow} needs less steps to achieve the same image quality. Latent sampling uses 4 steps and the midpoint solver, which takes $\approx 5 \mathrm{ms}$ whilst the decoder $\theta_D$ takes $\approx 3 \mathrm{ms}$. The reason inference in pixel space (RGB) is faster compared to the \gls{sdv} compressed image is the high decoding and inference cost of the \gls{sdv} decoder model ($\approx 20 \frac{\mathrm{ms}}{\mathrm{img}}$). The tests are done using unconditional samples for all models and \cref{fig:generation_unconditional} visualizes 32 random images produced with our method from each dataset. 

An interesting phenomenon occurs with \gls{lediflow} (\gls{sdv}) that the \gls{cmmd} score does not decrease with more steps. We have also seen this behavior using the 3rd order Heun method and examined a set of images with deterministic seeding (see Appendix). The images look visually very similar and only differ in minor details, which get picked up by the \gls{cmmd} metric~\cite{Jayasumana2023rethinking}. Findings on \gls{ode} dynamics \cite{karras2022edm,lee2023minimizing} describe, how small errors in the vector field prediction can accumulate to a worse result.

\subsection{Conditional Applications}
\label{subsec:results-cond_appl}
In this section, we look at applications of our method when modifying the latent space vector $z$. We start with a popular application when using \glspl{vae}, interpolating between points in latent space. In contrast to normal \glspl{vae}, we have two possibilities of doing the interpolation, linear in $Z$ or linear in $\mathcal{N}(0, 1)$ using $\theta_L$ for the transformation.

\begin{figure}
    \centering
    \includegraphics[width=0.17\linewidth]{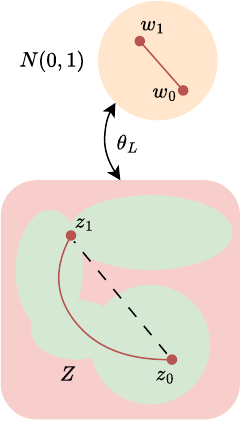}
    \includegraphics[width=0.65\linewidth]{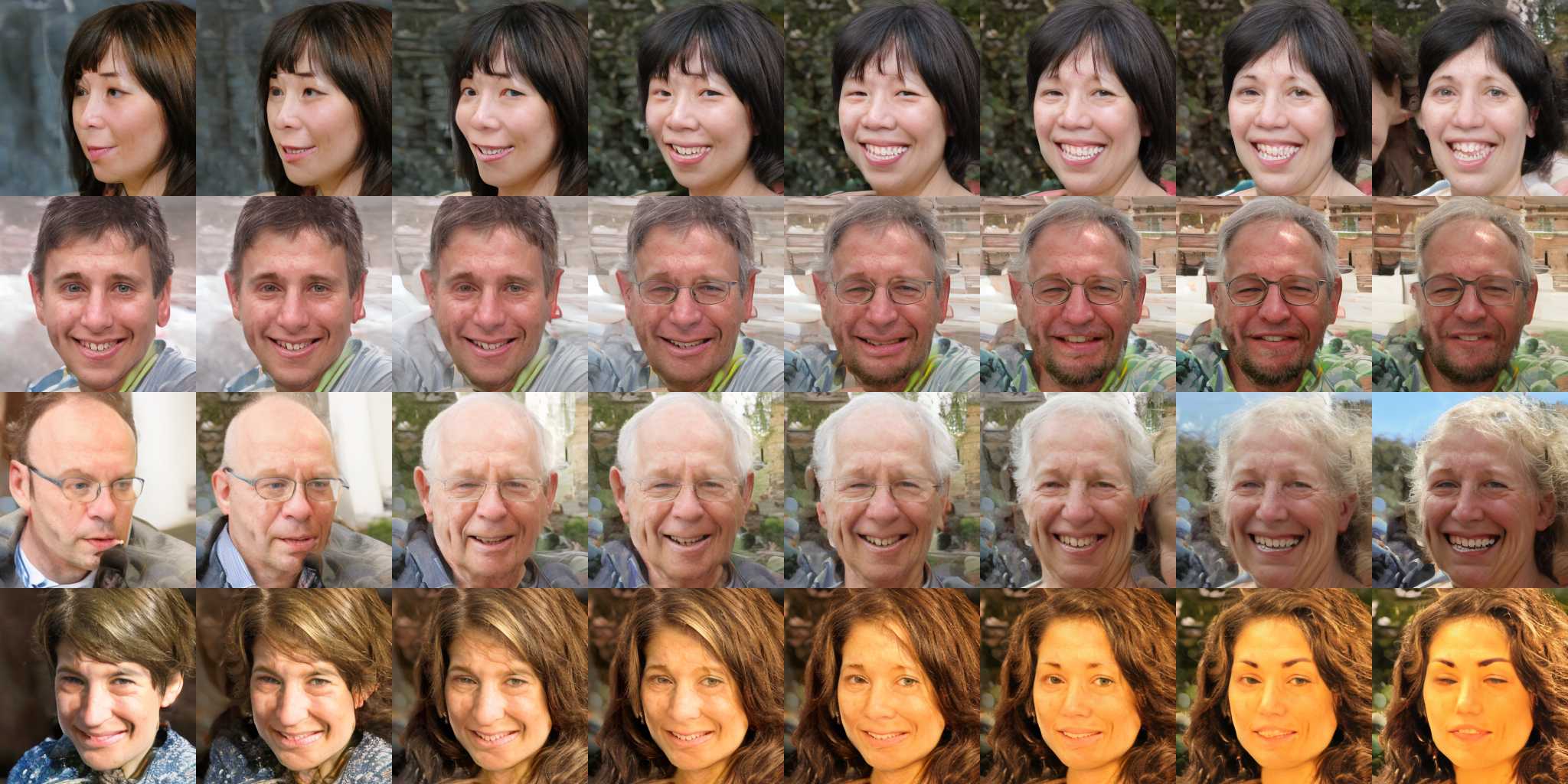}
    \caption{(Left) Visualization of the interpolation of samples $w \in \mathcal{N}(0,1)$ that are transformed into valid samples $z \in Z$ using $\theta_L$. (Right) Generated images from random latent vectors $z_0$ on the left and $z_1$ on the right using a constant noise seed.}
    \label{fig:ffhq_interpolation}
\end{figure}

In \cref{fig:ffhq_interpolation} on the left, we see problems that can occur when using the first approach. We do not know the high-dimensional shape of $Z$ and where most of the data is located. The shape might be nonconvex, thus, interpolating linearly can cross sparse or empty regions. To avoid these regions, we use method two and linearly interpolate in $\mathcal{N}(0,1)$ between two reverse-mapped vectors $w_0$ and $w_1$ given by reverse flow matching using $\theta_L$. The interpolated value is then mapped back to $Z$ using the latent sampler $\theta_L$ from \cref{subsec:method-lat_samp}, which results in a non-linear interpolation between $z_0$ and $z_1$. Looking at image samples on the right, we see that in all cases, the interpolated image contains valid data although with varying quality.

\begin{figure}
    \centering
    \includegraphics[width=0.49\linewidth]{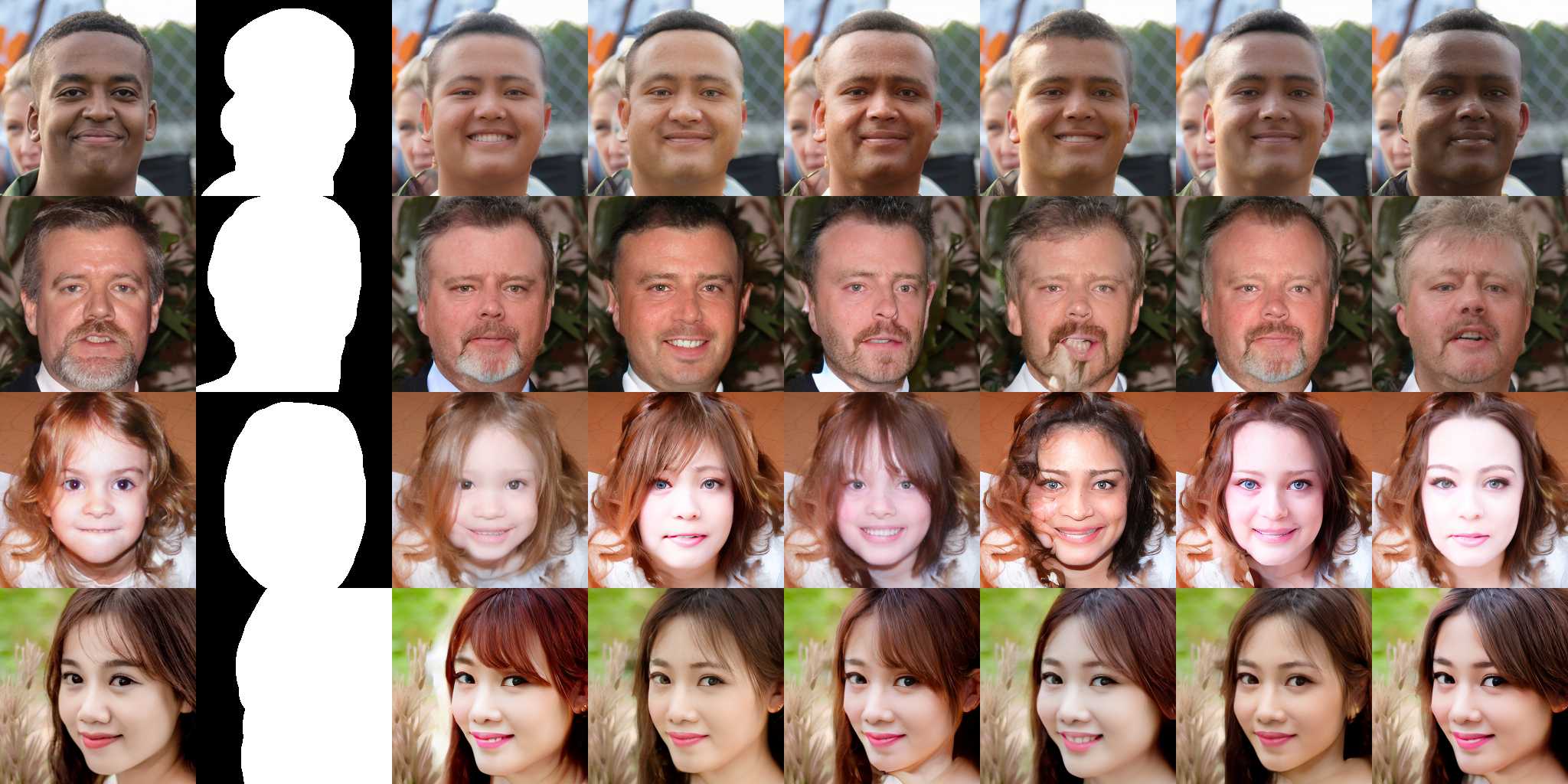}
    \includegraphics[width=0.49\linewidth]{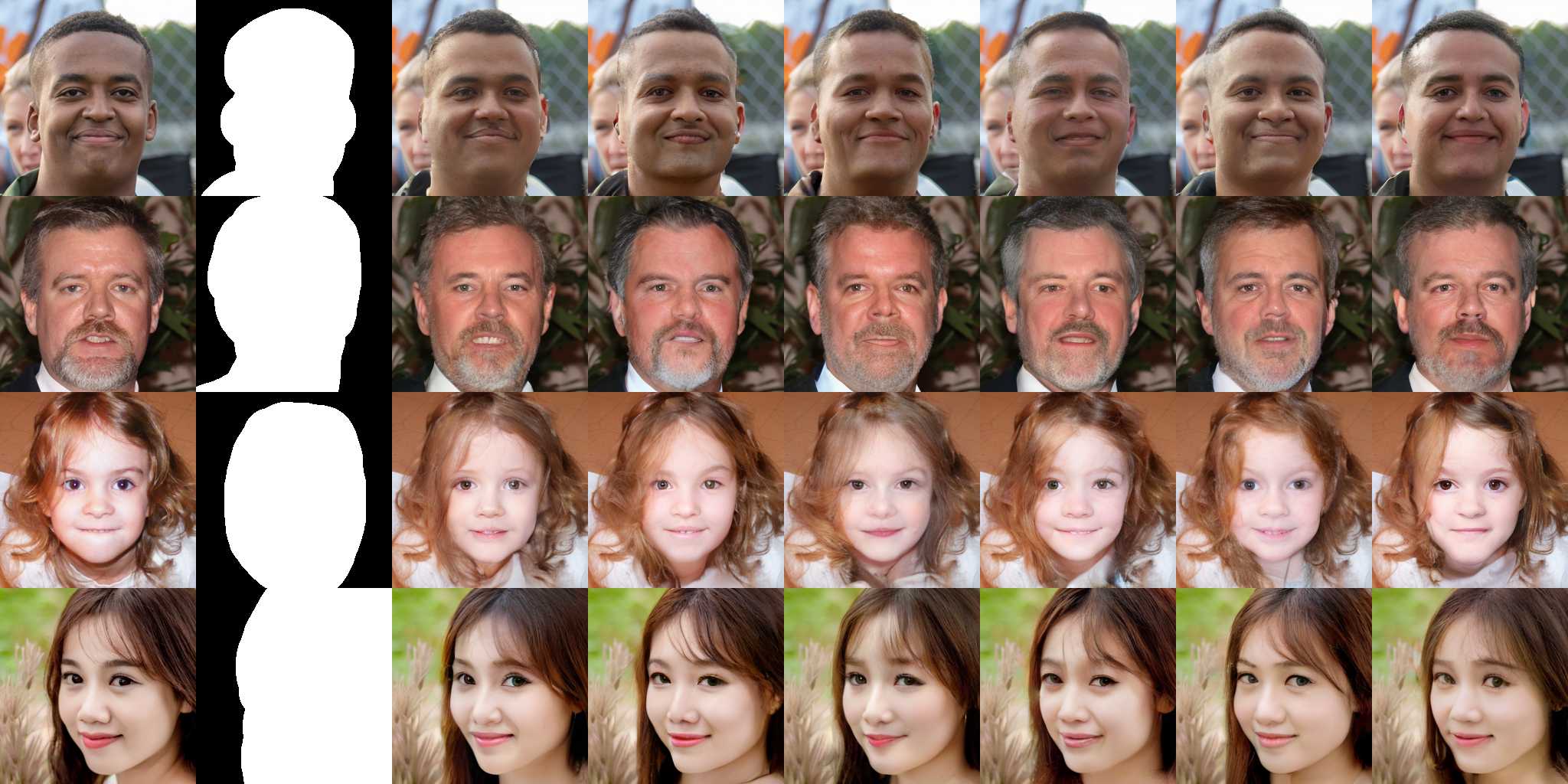}
    \caption{Our method is used to inpaint the original image (1st column) using a binary mask (2nd column). On the left half, we vary the latent space vector $z$ with a constant noise seed for drawing $x \sim P_L$ and on the right half, $z$ is constant and we vary the seed for drawing $x \sim P_L$.}
    \label{fig:ffhq_ano}
\end{figure}

Another popular example is image inpainting. We show this task briefly in \cref{fig:ffhq_ano} where the input image $y$ (left) is passed through the full auxiliary model pipeline to get a latent vector $z$ and an initial noisy image $x \sim P_L$. We also provide a mask $m$ indicating where the \gls{fmm} should generate pixels. The generation is already given by the general \gls{fm} formulation. The vector prediction of $\theta_\mathrm{FM}$ is used to calculate the flow of missing pixels provided by $m$. The modified vector field $\hat v : [0,1] \times \mathbb{R}^d \rightarrow \mathbb{R}^d$ for inpainting is given by:

\begin{align}
    \hat v_t(\phi_t(x)) &= 
    \begin{cases}
        v_t(\phi_t(x);\theta_\mathrm{FM}) & \text{where } m == 1\\
        y - x & \text{else}
    \end{cases}.
\end{align}

Using this equation, the flow matching process is forced to move in a straight line for every non-masked pixel towards $y$.

On the left half in \cref{fig:ffhq_ano}, we mapped $z \xrightarrow{\theta_L} w \in \mathcal{N}(0,1)$ using $\theta_L$ and reverse \gls{ode} solving and added a small random vector $r \in \mathcal{N}(0,1)$ to $w$ and then transformed it back using $w + \alpha r \xrightarrow{\theta_L} \hat z$ with $\alpha = 0.5$. This alters the latent vector and, consequently, the identity of the face slightly. We see that all non-masked areas do not change compared to the original image, whilst the masked areas are seamlessly inpainted using our method. On the right half, we see that changing the noise with a constant vector $z$ results in small details changing, while the overall face style is preserved.

More results and comparisons (i.e. plots, images, \gls{fid} values) are provided in the Appendix.

\section{Limitations and Broader Impacts}
\label{subsec:results-limits}
First, our method is highly dependent on the quality of the learned prior. A bad prior might be worse than using a Gaussian prior, as the transformation map might be more complex. Second, while \gls{lediflow} demonstrates effectiveness on domain-specific datasets like \gls{ffhq}, \gls{afhq}, \gls{lhq} the current model configuration (based on the \gls{hdit} S/4 backbone) struggles with the complexity and diversity of large-scale datasets such as ImageNet~\cite{deng2009imagenet}. Preliminary experiments indicated that neither \gls{lediflow} nor the baseline \gls{fm} achieves satisfactory performance on ImageNet at this scale, suggesting that significant architecture scaling or adaptation is required for such challenging domains. Third, a bad variance prediction $\sigma(z;\theta_D) \approx 0$ can lead to a mostly discrete prior, which is challenging to use for \gls{fm}. Fourth and finally, this paper has only shown a single model architecture, which is used in all three of the presented steps \gls{lediflow} needs to generate an image. Optimized architectures for the different steps can lead to better performance overall.

\glspl{lediflow} efficiency improvements can enhance accessibility to creative tools and support research, for example, through synthetic data generation or applications like face anonymization demonstrated in this work. However, advancements in realistic image synthesis also pose risks, such as potential misuse for disinformation and the amplification of data biases (e.g., from \gls{ffhq}, \gls{afhq} datasets). Continued research into content detection and bias mitigation remains crucial for responsible development. To promote responsible use of our contributions, our released code will be available under an MIT license. Pretrained models will be accompanied by Model Cards detailing their capabilities, limitations, ethical considerations, and data sources.

\section{Conclusion and Future Work}
\label{sec:conc_fw}

This work presents \gls{lediflow}, a scalable method that utilizes a learned prior distribution to enhance the inference performance of flow matching models. Instead of relying on a static Gaussian distribution as the prior, \gls{lediflow} uses a learned distribution for the initial step of the \gls{fm} algorithm. This approach reduces computational requirements, needing only half to one-third of the inference time compared to baseline \gls{fm}, while achieving competitive, or superior image quality, particularly on datasets like \gls{ffhq}(\gls{sdv}) and \gls{lhq}. We prove the capabilities of \gls{lediflow} by evaluating unconditional generation capabilities on multiple datasets and comparing against the reference \gls{fm} implementation. Our work primarily targets smaller models for specific use cases that involve images with a similar context, such as faces~\cite{Karras2019ffhq}, animals~\cite{ALIS} or landscapes~\cite{choi2020stargan}. Additionally, we also examine image inpainting and latent space interpolation, two popular applications of image generation models showing promising results. 
Future work could focus on applying different architectures on the components from \gls{lediflow} to improve the quality while maintaining affordable inference time. A potential future application of full-text-to-image generation would also be an interesting test case. This could demonstrate the general scalability of \gls{lediflow} and how the prediction of the prior responds to a more diverse dataset comprising various scenes.

\begin{ack}
    This work is supported by the Helmholtz Association Initiative and Networking Fund on the HAICORE@KIT partition. The authors gratefully acknowledge the computing time provided on the high-performance computer HoreKa by the National High-Performance Computing Center at KIT (NHR@KIT). This center is jointly supported by the Federal Ministry of Education and Research and the Ministry of Science, Research and the Arts of Baden-Württemberg, as part of the National High-Performance Computing (NHR) joint funding program (https://www.nhr-verein.de/en/our-partners). HoreKa is partly funded by the German Research Foundation (DFG). The Helmholtz Association funds the authors N.~Friederich and R.~Mikut under the program "Natural, Artificial and Cognitive Information Processing (NACIP)" and through the graduate school "Helmholtz Information \& Data Science School for Health (HIDSS4Health)". This paper emerged during the research project \textit{ANYMOS - Competence Cluster Anonymization for networked mobility systems} and was funded by the German Federal Ministry of Education and Research (BMBF) as part of \textit{NextGenerationEU} of the European Union.
\end{ack}

\bibliographystyle{abbrvnat}
\bibliography{sample-base}

\newpage
\printglossaries

\newpage
\section*{Appendix}

\begin{table}[h]
    \centering
    \begin{tabular}{c|c}
        Symbol & Description\\
        \hline
        $P$ & Source distribution\\
        $P_L$ & Learned prior distribution\\
        $P_N$ & Gaussian normal distribution $\mathcal{N}(0,1)$\\
        $Q$ & Target distribution (i.e. images)\\
        $Z$ & Latent space distribution induced by the VAE model\\
        \hline
        $x \sim P$ & Source sample from $P$\\
        $y \sim Q$ & Target sample from $Q$\\
        $t \in [0,1]$ & Timestep\\
        $v_t(\phi_t(x))$ & Learned vector field for sample $x$ time-dependent on $t$\\
        $v_t(\phi_t(x); \theta)$ & Learned vector field for sample $x$ time-dependent on $t$ using the neural network $\theta$\\
        $\phi_t(x)$ & Transformation state at timestep $t$ when initialized with $x$\\
        $\hat y$ & Result after solving the flow matching integral $\hat y = \phi_0(x) + \int_0^1 v_t(\phi_t(x)) dt$\\
        $z \in Z$ & Latent vector for an image sample $y$\\
        $\mu(y;\theta_D)$ & Mean of a latent variable\\
        $\sigma(y;\theta_D)$ & Standard deviation of a latent variable\\
        $\mu(z;\theta_E)$ & Mean of the a prior variable inducing $P_L$\\
        $\sigma(z;\theta_E)$ & Standard deviation of the a prior variable inducing $P_L$\\
        \hline
        $\theta_\mathrm{FM}$ & Flow matching model\\
        $\theta_E$ & VAE Encoder model\\
        $\theta_D$ & VAE Decoder model\\
        $\theta_L$ & Latent sampling model\\
        \hline
    \end{tabular}
    \caption{Math symbols used in the paper}
\end{table}

\begin{table*}[h]
    \centering
    \caption{Parameters used for our different models}
    \begin{tabular}{ccc}
        \hline
        Image Type & RGB & SDV \\
        Resolution & $256^2$ & $32^2$ \\
        \hline
        \textbf{Flow Matching Model} & \\
        \# Parameters & 160M & 154M \\
        Patch Size & 4 & 1 \\
        Levels (Local + Global Attention) & 2 + 1 & 1 + 1 \\
        Depth & [2, 2, 8] & [2, 8] \\
        Widths & [256, 512, 1024] & [512, 1024] \\
        Attention Head Dim & 64 & 64 \\
        Mapping Depth & 2 & 2 \\
        Mapping Width & 768 & 768 \\
        Neighborhood Kernel Size & 7 & 7 \\
        Batch Size & 128 & 64\\
        \hline
        \textbf{Auxiliary model}\\
        \# Parameters & 37M & 35M \\
        Levels (Local + Global Attention) & 4 + 1 & 3 + 1 \\
        Depth & [2, 2, 2, 2, 2] & [2, 2, 2, 2] \\
        Widths & [128, 256, 512, 512] & [256, 512, 512] \\
        Attention Head Dim & 16 & 16 \\
        Latent Shape & $8 \times 8 \times 4$ & $8 \times 8 \times 4$ \\
        Neighborhood Kernel Size & 7 & 7 \\
        Batch Size & 256 & 128\\
        \hline
        \textbf{Latent Sampler}\\
        \# Parameters & 17M & 17M \\
        Levels (Global Attention) & 1 & 1 \\
        Depth & [8] & [8] \\
        Widths & [384] & [384] \\
        Attention Head Dim & 64 & 64 \\
        Batch Size & 512 & 256\\
        \hline
        \textbf{Training Settings}\\
        Optimizer & AdamW & AdamW\\
        Betas & (0.9, 0.99) & (0.9, 0.99)\\
        Learning Rate & 3e-4 & 3e-4\\
        Precision & bfloat16 & bfloat16\\
        \hline
    \end{tabular}
    \label{tab:model_table}
\end{table*}

\begin{figure}[h]
    \centering
    \includegraphics[width=0.32\linewidth]{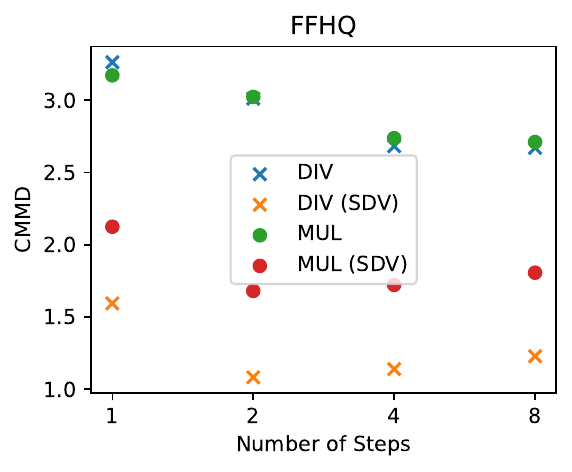}
    \includegraphics[width=0.32\linewidth]{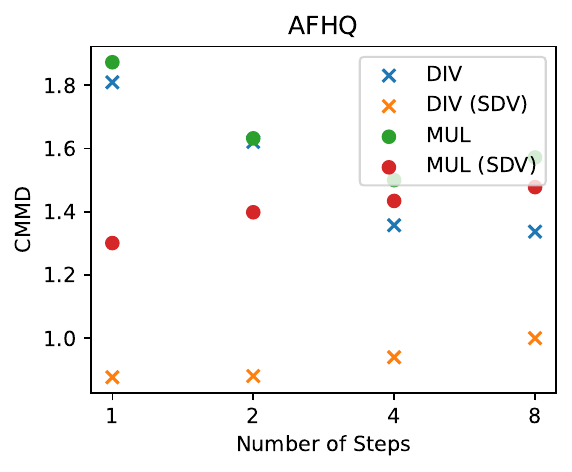}
    \includegraphics[width=0.32\linewidth]{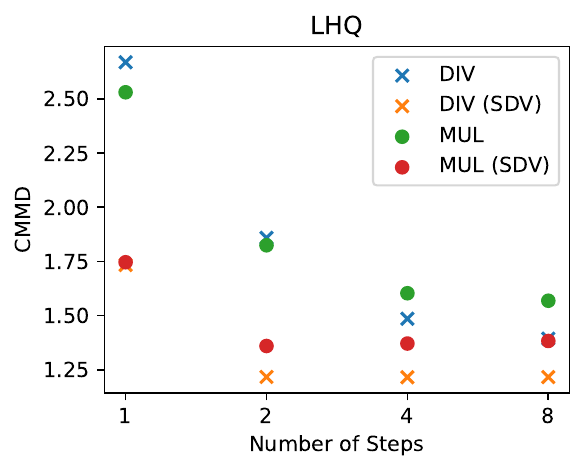}
    \caption{Comparison of using multiplication vs division importance scaling in \cref{eq:flow_loss_importance}.}
    \label{fig:cmmd_mul_div}
    \vspace{-10pt}
\end{figure}

\begin{figure}[h]
    \centering
    \includegraphics[width=0.32\linewidth]{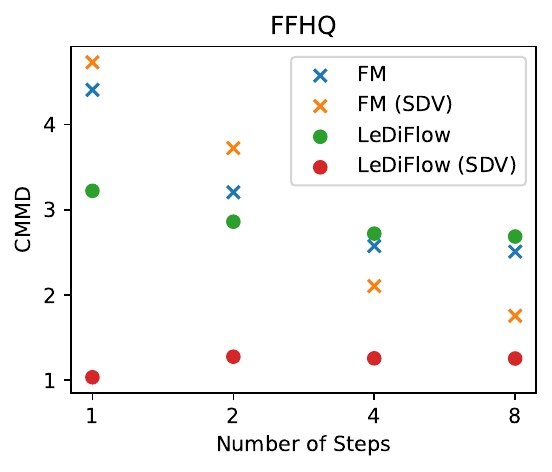}
    \includegraphics[width=0.32\linewidth]{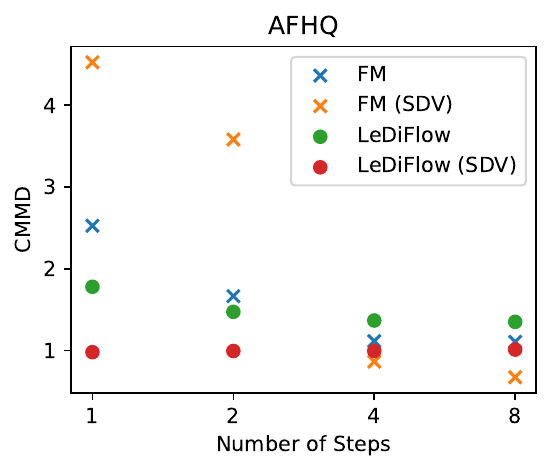}
    \includegraphics[width=0.32\linewidth]{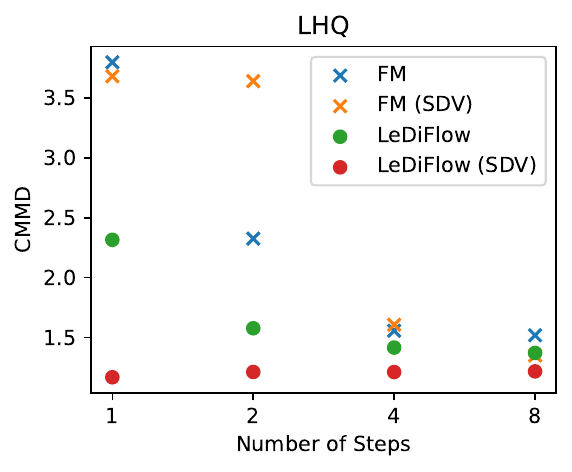}
    \caption{Comparison of \gls{fm} and \gls{lediflow} similar to \cref{fig:inf_performance}.}
    \label{fig:cmmd_heun3}
    \vspace{-10pt}
\end{figure}

\begin{figure}[h]
    \centering
    \includegraphics[width=0.32\linewidth]{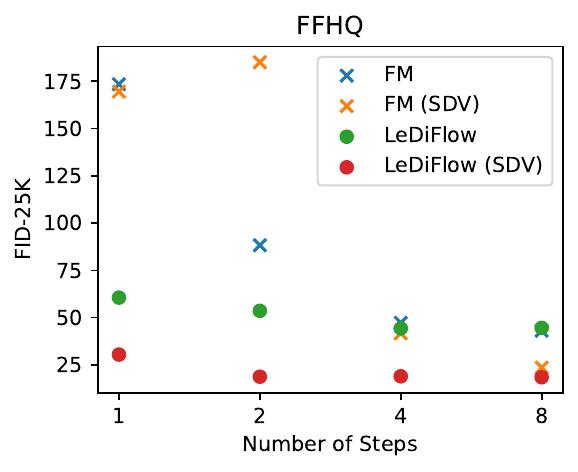}
    \includegraphics[width=0.32\linewidth]{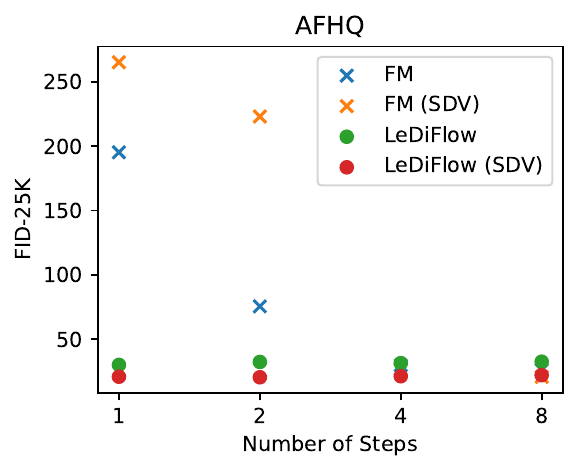}
    \includegraphics[width=0.32\linewidth]{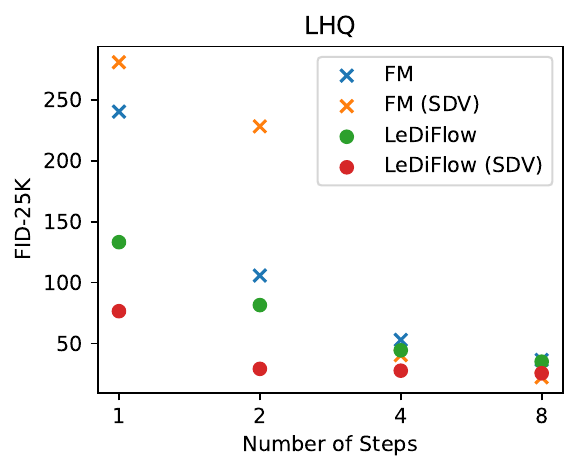}
    \caption{\gls{fid} values comparing \gls{fm} with \gls{lediflow} similar to \cref{fig:inf_performance}.}
    \label{fig:fid}
    \vspace{-10pt}
\end{figure}

\begin{figure}[h]
    \centering
    \includegraphics[width=1.0\linewidth]{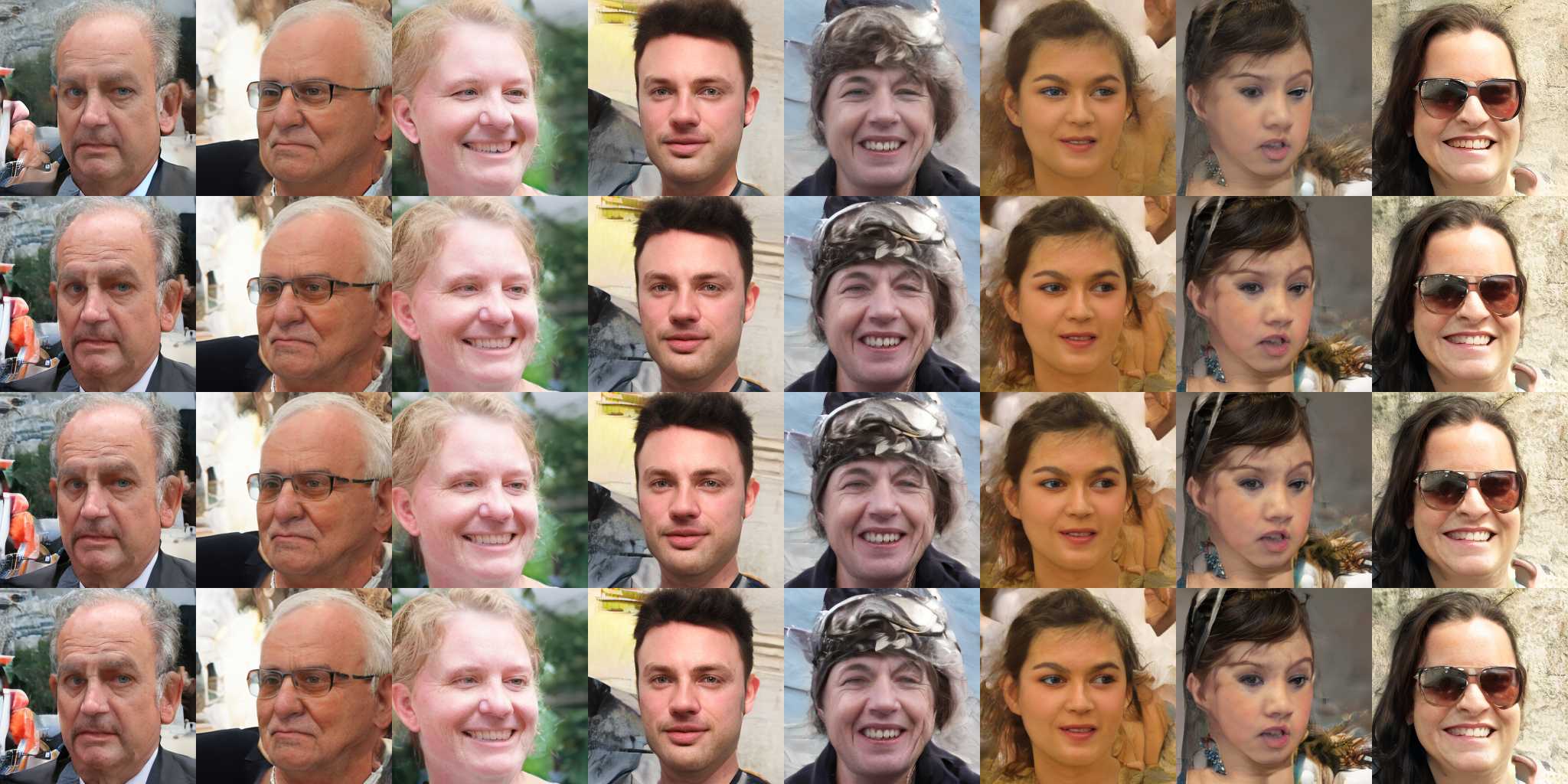}
    \caption{From top to bottom: 1, 2, 4 and 8 midpoint steps of inference using \gls{lediflow} on \gls{ffhq}~\cite{Karras2019ffhq}.}
    \label{fig:cmmd_qual}
    \vspace{-10pt}
\end{figure}

\begin{figure}[h]
    \centering
    \begin{subfigure}[b]{0.44\linewidth}
        \centering
        \includegraphics[width=\linewidth]{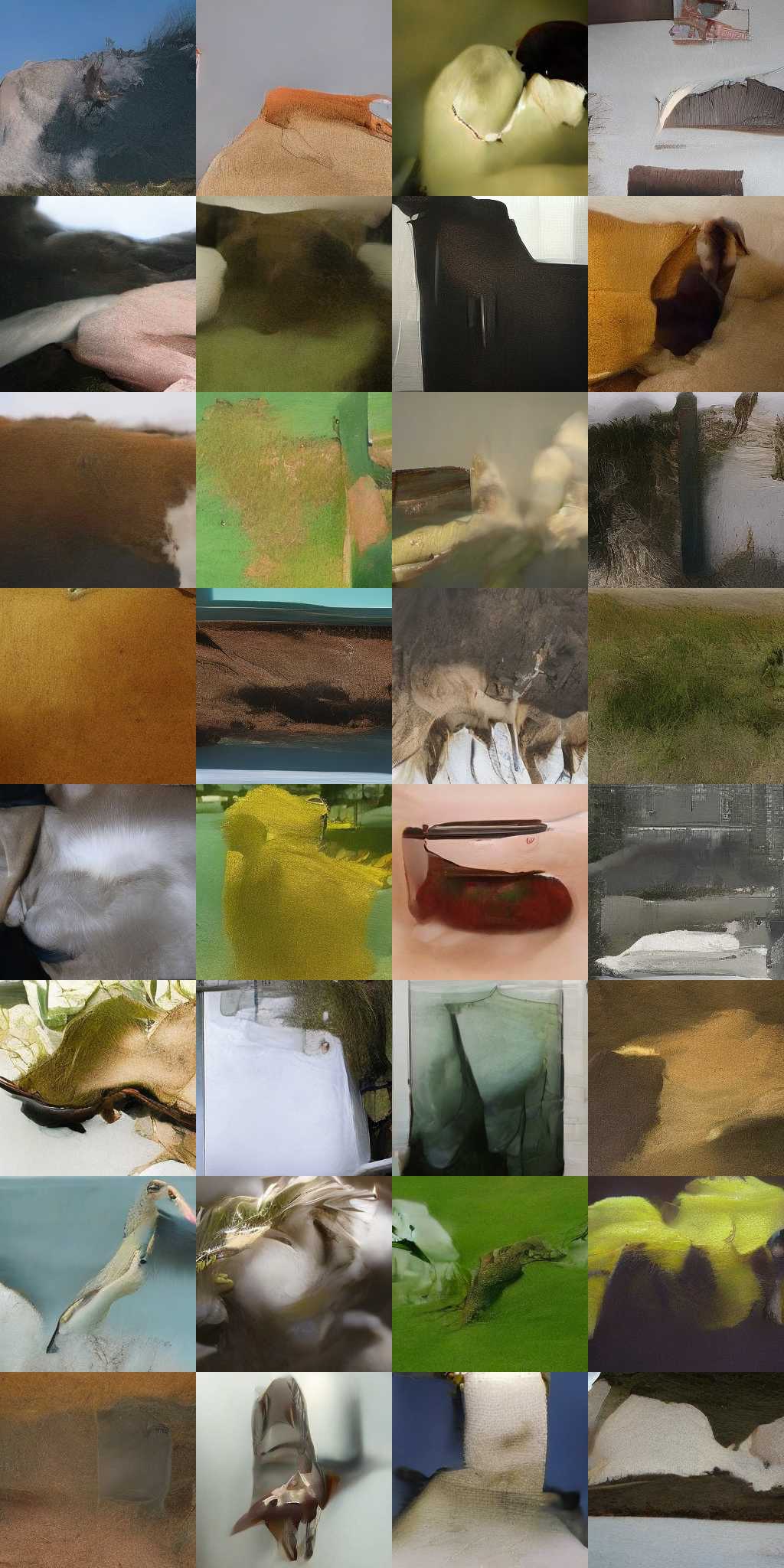}
        \caption{\gls{lediflow}}
    \end{subfigure}
    \hfill
    \begin{subfigure}[b]{0.44\linewidth}
        \centering
        \includegraphics[width=\linewidth]{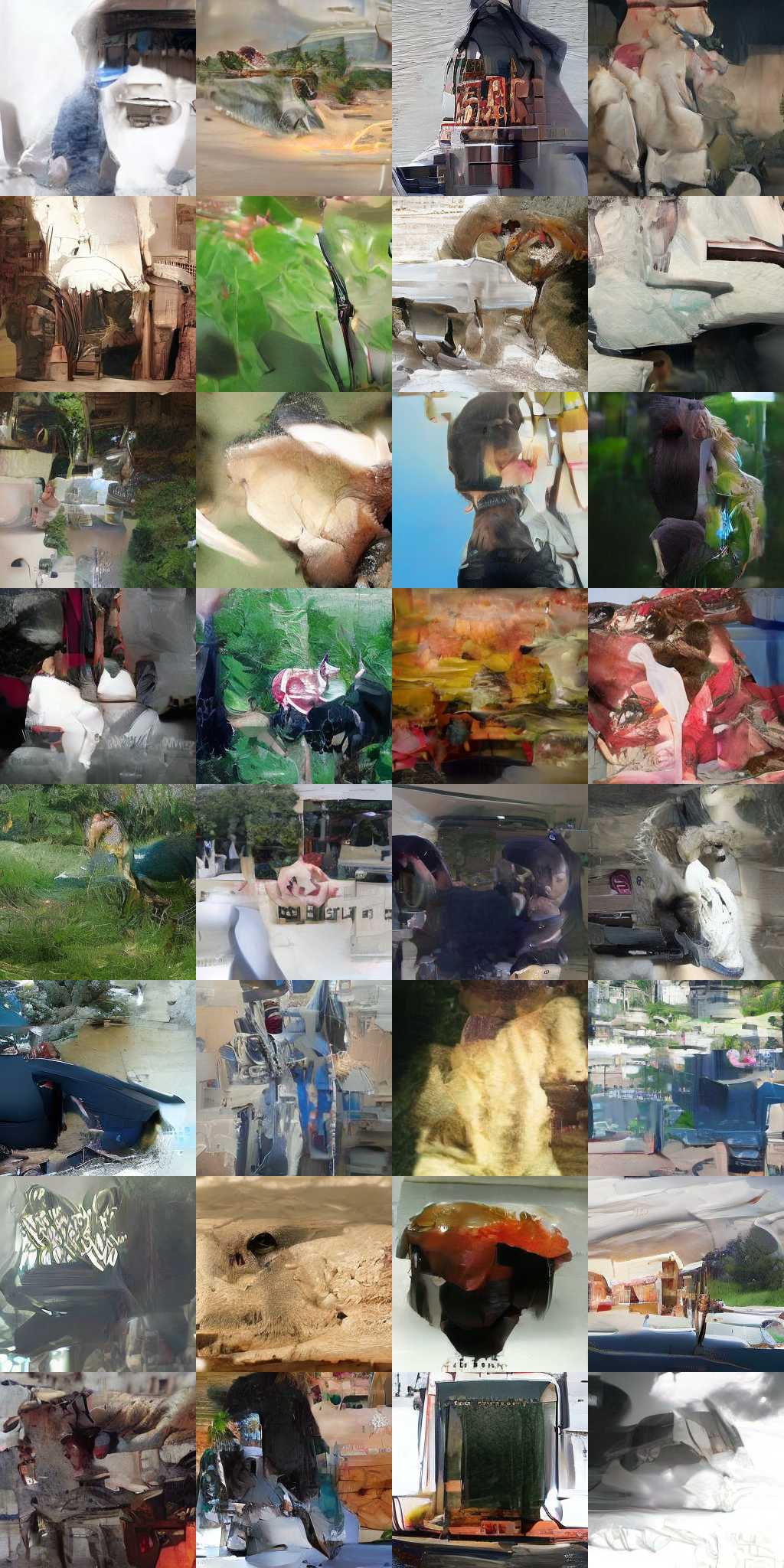}
        \caption{\gls{fm}}
    \end{subfigure}
    \caption{ImageNet \cite{deng2009imagenet} results using \gls{lediflow} and \gls{fm}.}
    \label{fig:imagenet}
    \vspace{-10pt}
\end{figure}

\begin{figure}[h]
    \centering
    \begin{subfigure}[b]{1\linewidth}
        \centering
        \includegraphics[width=\linewidth]{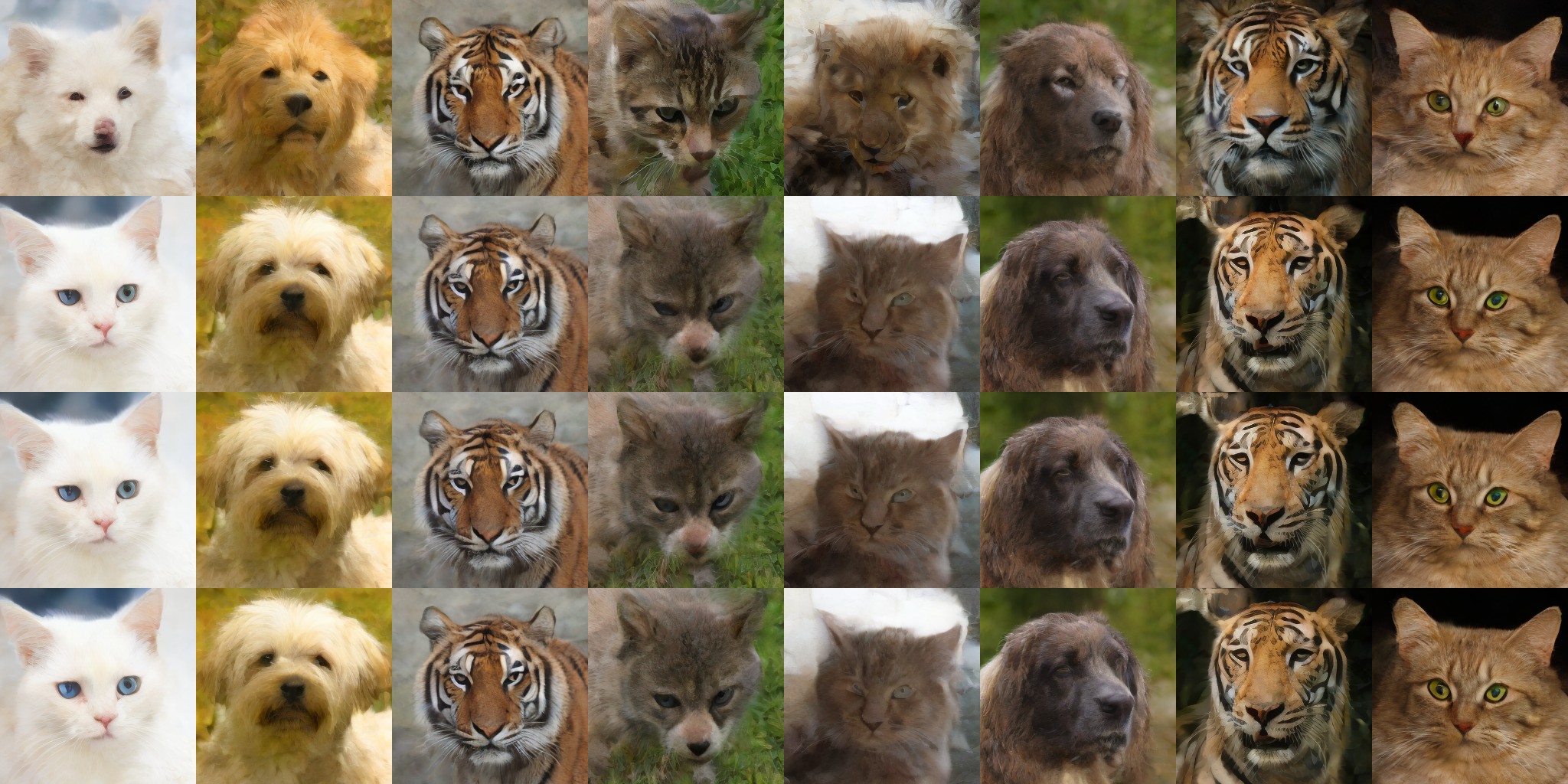}
        \caption{\gls{afhq}}
    \end{subfigure}
    
    \vspace{10pt} 

    \begin{subfigure}[b]{1\linewidth}
        \centering
        \includegraphics[width=\linewidth]{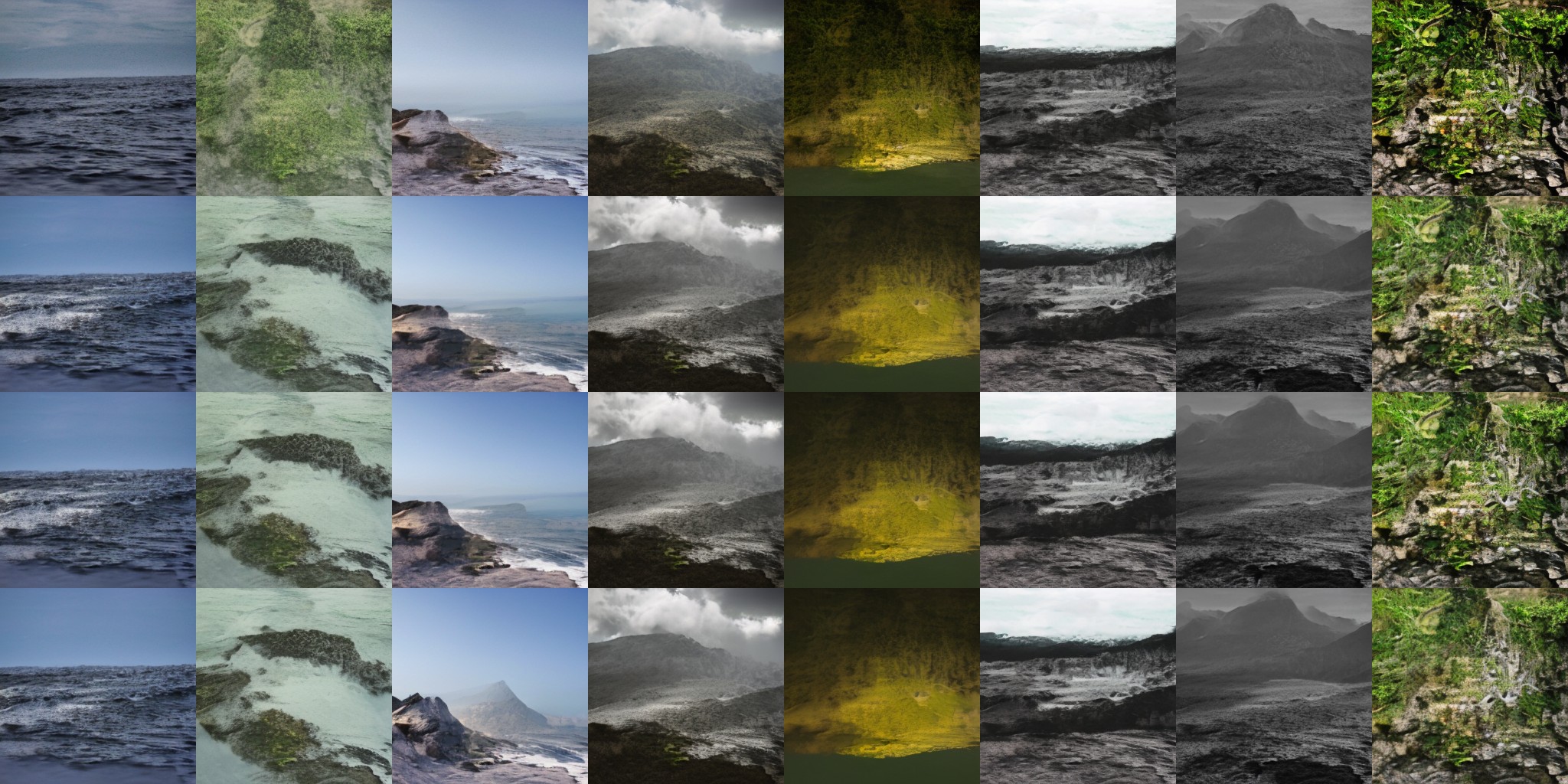}
        \caption{\gls{lhq}}
    \end{subfigure}
    
    \caption{Images generated with varying number of steps of the latent sampler. From top to bottom: 1, 2, 4, and 8 steps.}
    \label{fig:qual_lat}
    \vspace{-10pt}
\end{figure}
\begin{table*}[h]
    \centering
    \caption{Licenses of Assets used in this work}
    \begin{tabular}{ccc}
        \hline
        Asset & License\\
        \gls{ffhq} \cite{Karras2019ffhq} & CC-BY-NC-SA 4.0\\
        \gls{afhq} \cite{ALIS} & CC-BY-NC 4.0\\
        \gls{lhq} \cite{choi2020stargan} & CC BY 2.0\\
        ImageNet \cite{deng2009imagenet} & Custom Non Commercial\\
        \gls{sdv} \cite{stabilityai2024stable} & CreativeML Open RAIL++-M License\\
        \hline
    \end{tabular}
    \label{tab:licenses}
\end{table*}

\end{document}